\pdfoutput=1

\documentclass[11pt]{article}

\usepackage[preprint]{acl}

\usepackage{times}
\usepackage{latexsym}

\usepackage[T1]{fontenc}

\usepackage[utf8]{inputenc}

\usepackage{microtype}

\usepackage{inconsolata}

\usepackage{graphicx}

\usepackage{booktabs}
\usepackage{multirow}
\usepackage{xspace}
\usepackage{array}
\usepackage{amsmath} 
\usepackage{amsfonts}
\usepackage{makecell}

\newcommand{\ours}{\textsc{SeaLong}\xspace}

\title{Large Language Models Can Self-Improve in Long-context Reasoning}

\author{
 \textbf{Siheng Li$^{\heartsuit}$}
 \quad
 \textbf{Cheng Yang$^{\heartsuit}$}
 \quad
 \textbf{Zesen Cheng$^{\spadesuit}$}
 \quad
 \textbf{Lemao Liu$^{\diamondsuit}$}
 \quad
 \textbf{Mo Yu$^{\diamondsuit}$}
 \\
 \quad
 \textbf{Yujiu Yang$^{\clubsuit}$}
 \quad
 \textbf{Wai Lam$^{\heartsuit}$}
\\ 
 $^{\heartsuit}$The Chinese University of Hong Kong
 \\
 \quad
 $^{\spadesuit}$Peking University
 \quad
 $^{\clubsuit}$Tsinghua University
 \quad
 $^{\diamondsuit}$Tencent
\\
\small{\href{mailto:sihengli24@gmail.com}{sihengli24@gmail.com}} 
\\
 \small{
   \textbf{Correspondence:} \href{mailto:moyumyu@tencent.com}{moyumyu@tencent.com} \quad
   \href{mailto:wlam@se.cuhk.edu.hk}{wlam@se.cuhk.edu.hk}
 }
}

\begin{document}
\maketitle

\begin{abstract}
    Large language models (LLMs) have achieved substantial progress in processing long contexts but still struggle with long-context reasoning. Existing approaches typically involve fine-tuning LLMs with synthetic data, which depends on annotations from human experts or advanced models like GPT-4, thus restricting further advancements. To address this issue, we investigate the potential for LLMs to self-improve in long-context reasoning and propose \ours, an approach specifically designed for this purpose. This approach is straightforward: we sample multiple outputs for each question, score them with Minimum Bayes Risk, and then apply supervised fine-tuning or preference optimization based on these outputs. Extensive experiments on several leading LLMs demonstrate the effectiveness of \ours, with an absolute improvement of $4.2$ points for Llama-3.1-8B-Instruct. Furthermore, \ours achieves superior performance compared to prior approaches that depend on data produced by human experts or advanced models. We anticipate that this work will open new avenues for self-improvement techniques in long-context scenarios, which are essential for the continual advancement of LLMs.\footnote{The repository can be accessed at \url{https://github.com/SihengLi99/SEALONG}.}
\end{abstract}
\section{Introduction}

\begin{table*}[htbp]
\centering
\resizebox{\linewidth}{!}{
\begin{tabular}{lcccccc}
\toprule
    \multirow{2}{*}{Prompt}                                        & \multicolumn{3}{c}{Llama-3.1-8B-Instruct}                              & \multicolumn{3}{c}{Llama-3.1-70B-Instruct}      \\
                                                                     \cmidrule(lr){2-4}                                                       \cmidrule(lr){5-7}                                 
                                                                   & HotpotQA              & MuSiQue                  & 2WikiMQA            & HotpotQA              & MuSiQue                  & 2WikiMQA       \\
    \midrule    
    Default                                                        & 55.5                  & 33.0                     & 66.0                & 60.0                  & 54.0                     & 77.0           \\    
    Direct answer                                                  & 49.0                  & 28.5                     & 55.0                & 61.5                  & 51.5                     & 74.0           \\       
    Think step-by-step \citep{kojima2022large}                     & 62.5                  & \textbf{50.5}            & 77.5                & 75.5                  & 62.5                     & 85.0           \\
    Fact-and-reflection \citep{zhao-etal-2024-fact}                & \textbf{67.0}         & 49.0                     & 76.5                & \textbf{78.0}         & 62.0                     & 84.0           \\ 
    Plan-and-solve \citep{wang2023plan}                            & 64.0                  & 49.5                     & \textbf{82.0}       & 74.0                  & \textbf{68.5}            & \textbf{85.5}  \\
\bottomrule
\end{tabular}
}
\caption{
Comparison of various prompting methods. The best result is highlighted in bold.
}
\label{tab:prompting_methods}
\end{table*}

Large language models (LLMs) with long-context processing capabilities have spurred a range of novel applications, such as repository-level coding assistance \citep{jimenez2024swebench}, multi-document analysis \citep{wang2024leave} and autonomous agents \citep{ma2024agentboard}. Delivering high-quality service in these domains requires LLMs to \emph{reason effectively over long contexts}, necessitating the retrieval of essential details and integration of dispersed information throughout the reasoning process. While recent advancements have enabled LLMs to attain near-perfect accuracy on the needle-in-a-haystack (NIAH; \citet{needleinhaystack, li2024needlebench}) task \citep{hsieh2024ruler, yen2024helmet, dubey2024llama}, which involves locating evidence within vast amounts of irrelevant text, substantial performance declines persist on tasks that require reasoning over long contexts \citep{levy-etal-2024-task, hsieh2024ruler, vodrahalli2024michelangelo, yen2024helmet, li2024alr}, limiting their applicability in real-word scenarios.

To address this limitation, recent studies have investigated fine-tuning LLMs to improve long-context reasoning, with effective data synthesis as a primary challenge. Two main approaches have emerged: one relies on human annotations \citep{chen2024longlora, li2024making, li2024alr}, which are expensive and difficult to scale; the other leverages expert models, such as GPT-4o \citep{hurst2024gpt}, for data synthesis. For example, \citet{bai2024longalign, zhang2024extending, zhang2024longreward} apply self-instruct techniques \citep{wang2023self} to create long-context instruction-following data. Despite substantial progress, the dependence on pre-existing expert models limits the potential for achieving more advanced capabilities.

This work investigates whether LLMs can \emph{self-improve in long-context reasoning}. Drawing on evidence of LLMs' near-perfect long-context retrieval and strong reasoning abilities in general domains \citep{bubeck2023sparks, zhong2024evaluation}, we hypothesize that LLMs have untapped potential in long-context reasoning. Our preliminary studies show that refined prompting strategies achieve notable improvements over both default prompting methods and direct answer requests. Furthermore, scaling the number of sampled outputs reveals a marked performance gap between the optimal outputs and those derived via greedy search. These results suggest that LLMs hold substantial potential to advance in long-context reasoning.

Inspired by these observations, we propose a 
\underline{S}elf-improving method for r\underline{EA}soning over \underline{LONG}-contexts (\ours). This involves first sampling multiple reasoning trajectories from the LLM, then scoring each based on Minimum Bayes Risk (MBR) \citep{bickel1977mathematical}, which prioritizes outputs that are more consistent with others. This idea is intuitive, as reasoning trajectories that deviate from the majority are more likely to be hallucinations \citep{manakul2023selfcheckgpt, farquhar2024detecting}. Following this, we can either conduct supervised fine-tuning using high-scoring outputs or apply preference optimization by utilizing both high-scoring and low-scoring outputs.

We apply \ours to several leading LLMs and conduct evaluations on multiple long-context reasoning tasks \citep{bai2023longbench, yang2018hotpotqa, trivedi2022musique, ho2020constructing, dasigi2021dataset}. The results reveal that LLMs can self-improve in long-context reasoning. Specifically, \ours raises the score of Llama-3.1-8B-Instruct \citep{dubey2024llama} from $50.8$ to $55.0$. Additionally, \ours enables Qwen-2.5-14B-Instruct \citep{qwen2} to outperform its 32B variant ($54.7$ vs. $53.1$). In comparison to previous synthetic data, \ours demonstrate notable improvement without requiring human or expert model annotation. We hope that \ours can pave the way for self-improving approaches in long-context scenarios, supporting the continual advancement of LLM capabilities.

\section{Understanding the Potential of LLMs in Long-context Reasoning}
\label{sec:potential_of_llms_in_long_context_reasoning}

\begin{figure*}[t]
\centering
\includegraphics[width=\linewidth]{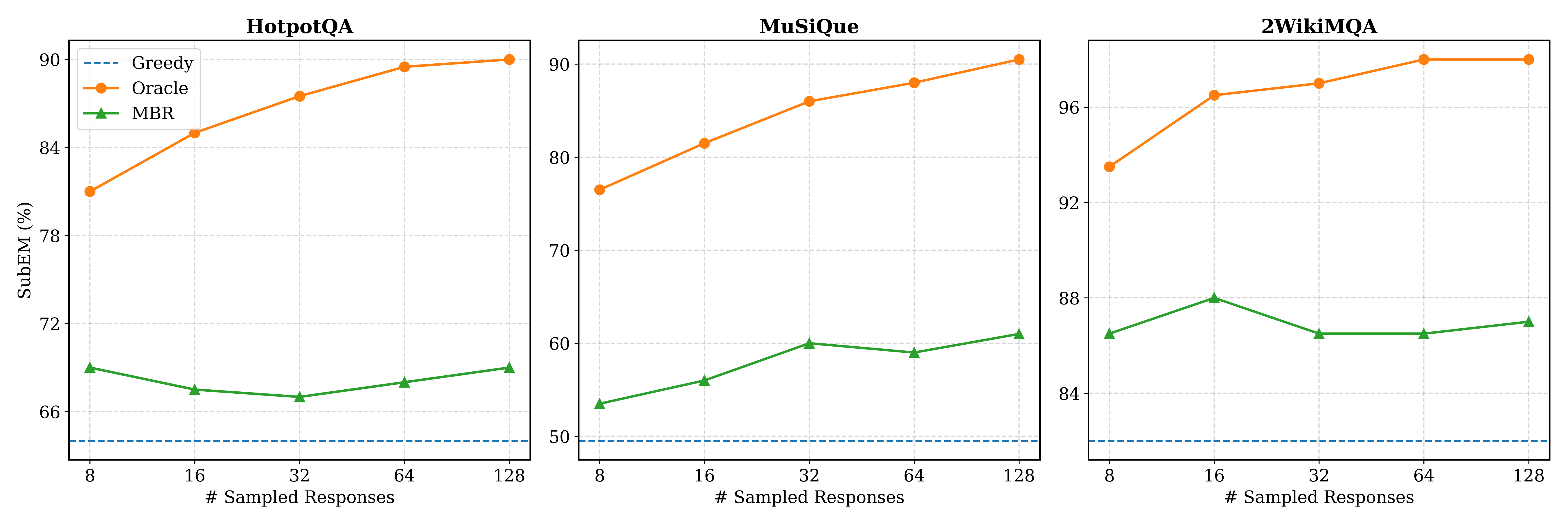}
\caption{
    Scaling up the number of sampled outputs improves the performance of both the oracle sample and MBR decoding (\textsection\ref{sec:self_supervision}). The results are based on Llama-3.1-8B-Instruct.
}
\label{figure:llms_potential_for_correct_long_context_reasoning}
\end{figure*}

We explore the potential of LLMs in long-context reasoning through experiments on three reasoning-intensive tasks from LongBench \citep{bai2023longbench}: HotpotQA \citep{yang2018hotpotqa}, MuSiQue \citep{trivedi2022musique} and 2WikiMQA \citep{ho2020constructing}. These tasks involve handling multiple documents within the context and addressing multi-hop questions that span several paragraphs. Following previous work \citep{mallen2023not, asai2024selfrag, yen2024helmet}, we use substring exact match (SubEM) for evaluation, assessing whether the golden answer is included in the output.

\subsection{Prompting Strategies Matter}
\label{sec:prompting_strategies_matter}
Numerous long-context evaluation benchmarks assess LLMs by simply asking them to respond to a query based on a long context \citep{bai2023longbench, an-etal-2024-l, zhang-etal-2024-bench, wang2024leave, yen2024helmet}. We suggest that this approach may underestimate LLMs' potential in long-context scenarios, particularly for questions requiring complex, multi-step reasoning to arrive at an answer. To further investigate this, we examine various prompting strategies for long-context reasoning, including:
\begin{itemize}
    \item \textbf{Default}: Prompting the LLM with the long context and a question.
    \item \textbf{Direct Answer}: Asking the LLM to directly answer the question based on the long context.
    \item \textbf{Think Step-by-step}: Providing the LLM with the context, question, and an instruction to think step-by-step \citep{kojima2022large}.
    \item \textbf{Fact-and-reflection}: Providing the LLM with the long context, question, and an instruction to first identify the relevant information from the long context, and then carry out step-by-step reasoning and provide the answer \citep{zhao-etal-2024-fact, li2024alr}.
    \item \textbf{Plan-and-solve}: Providing the LLM with the long context, question, and an instruction to first devise a plan and then follow it to solve the problem step-by-step \citep{wang2023plan}.
\end{itemize}
The detailed prompts for these strategies are presented in Tab. \ref{tab:prompts} (Appx. \ref{sec:prompts}). As shown in Tab. \ref{tab:prompting_methods}, prompting strategies play a crucial role in long-context reasoning. A notable performance gap exists between default prompting and reasoning-targeted prompting strategies, aligning with observations in short-context tasks \citep{wei2022chain, zhou2023leasttomost}. Manual inspection reveals that with an appropriate prompting strategy, the LLM breaks down multi-hop questions into simpler parts, addresses each part using the long context, and ultimately arrives at an answer.

\subsection{The Potential of LLMs for Correct Long-context Reasoning}
We further investigate the potential of LLMs for long-context reasoning by expanding the generation space. Specifically, we use temperature sampling to produce multiple outputs per question, evaluate each with SubEM, and designate the highest-scoring output as the oracle sample. As shown in Fig. \ref{figure:llms_potential_for_correct_long_context_reasoning}, there is a notable gap between oracle performance and that of greedy search, even with just $8$ outputs. Scaling up to $128$ samples achieves over $90\%$ correct answers. These results underscore the potential of LLMs for long-context reasoning and motivate the development of methods that enable LLMs to self-improve in this area.
\section{\ours}
\begin{figure*}[t]
\centering
\includegraphics[width=\linewidth]{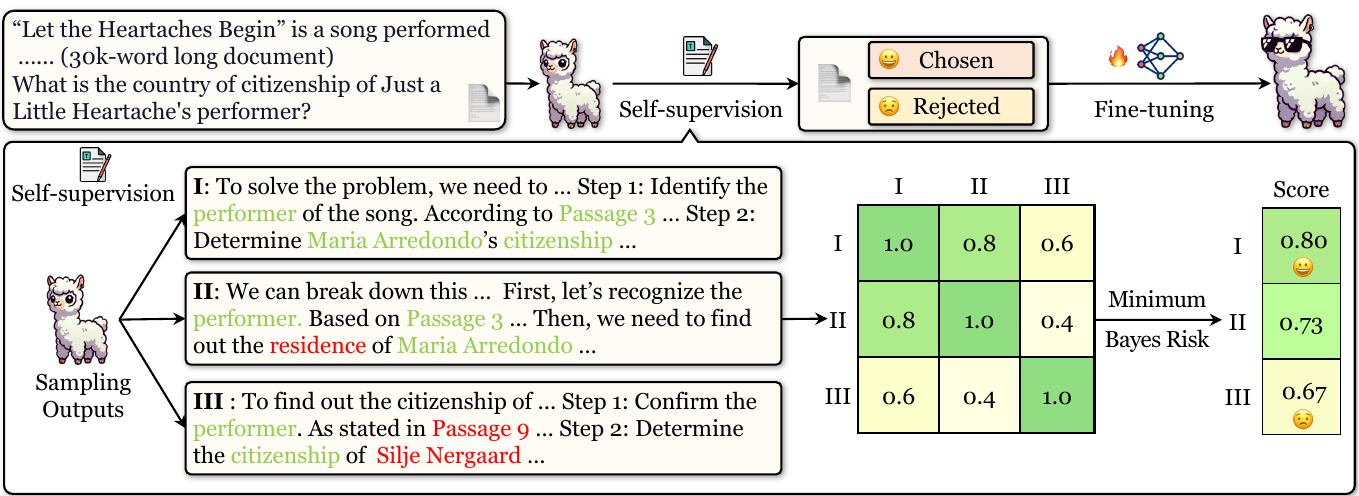}
\caption{
    \ours consists of two stages: self-supervision creation and fine-tuning. Given a long context and a corresponding query, multiple outputs are sampled, each assigned a score based on Minimum Bayes Risk. Fine-tuning is then conducted using either the highest-scoring output for supervised fine-tuning or both high-scoring and low-scoring outputs for preference optimization.
}
\label{fig:method}
\end{figure*}

Motivated by the potential of LLMs in long-context reasoning (\textsection\ref{sec:potential_of_llms_in_long_context_reasoning}), we propose \ours, a self-improving method for reasoning over long contexts. This approach consists of two stages: creating self-supervision and fine-tuning the model. An overview of \ours is provided in Fig. \ref{fig:method}.

\subsection{Self-supervision}
\label{sec:self_supervision}
We begin by leveraging plan-and-solve prompting \citep{wang2023plan} to sample multiple reasoning trajectories for each question and its corresponding long context. The primary challenge lies in evaluating these outputs. The fundamental idea behind \ours is that correct reasoning trajectories typically exhibit higher semantic consistency. For example, they tend to follow similar planning steps and reference the same information within the long context. This observation aligns with hallucination detection methods \citep{manakul2023selfcheckgpt, farquhar2024detecting}, where less consistent outputs are more likely to indicate hallucinations, representing incorrect reasoning in our scenario.

We formalize this idea using Minimum Bayes Risk (MBR) \citep{bickel1977mathematical, bertsch2023s, wu2024better}, which prioritizes outputs that exhibit higher consistency with others. In the MBR literature, the quality of an output is assessed by its expected utility under the model distribution \citep{bickel1977mathematical, kumar2004minimum, tromble2008lattice}:
\begin{equation}
    s(y) = \mathbb{E}_{y^* \sim \pi_\theta(y \mid x)} \left[ u(y, y^*) \right] \nonumber
\end{equation}
Here, $s(y)$ is the score assigned to output $y$, where $x$ denotes the input, including the long context, question and instruction. The term $\pi_\theta(y \mid x)$ represents the policy distribution of the LLM and the utility metric $u(y, y^*)$ assesses $y$ based on $y^*$. We approximate this expectation using the Monte Carlo method with $N$ sampled outputs:
\begin{equation}
    s(y) \approx \frac{1}{N} \sum_{i=1}^N \left[ u(y, y^*) \right] \nonumber
\end{equation}
The utility metric measures the consistency between two outputs. We use sentence embedding similarity as this metric, as it effectively captures the semantic alignment between the two reasoning trajectories. Formally:
\begin{equation}
    u(y, y^*) = \operatorname{Sim} \left( \operatorname{Emb}(y), \operatorname{Emb}(y^*) \right) \nonumber
\end{equation}

We employ a lightweight RoBERTa-based model \citep{liu2019roberta} to embed outputs and measure similarity with the inner product. This approach allows us to assign each output $y$ a score $s(y)$, and selecting the output with the highest score is referred to as MBR decoding \citep{bickel1977mathematical, bertsch2023s, wu2024better}. As demonstrated in Fig. \ref{figure:llms_potential_for_correct_long_context_reasoning}, MBR decoding substantially surpasses greedy search: with absolute improvements of $11.5\%$ on MuSiQue \citep{trivedi2022musique}, $5.0\%$ on HotpotQA \citep{yang2018hotpotqa}, and $5.0\%$ on 2WikiMultihopQA \citep{ho2020constructing} when $N=128$. These results highlight the potential for LLMs to self-improve by leveraging multiple samples and an effective evaluation metric based on output consensus, eliminating the need for human experts or advanced models. Furthermore, this evaluation approach produces preference pairs by contrasting high-scoring and low-scoring outputs, allowing straightforward preference optimization \citep{ouyang2022training, rafailov2024direct}.

\subsection{Fine-tuning} 
\begin{table*}[htbp]
\centering
\resizebox{\linewidth}{!}{
\begin{tabular}{lcccccc}
\toprule
    Model                                         & Qasper         & MultiFieldQA-En    & HotpotQA          & MuSiQue           & 2WikiMQA         & Avg.               \\
    \midrule
    Qwen-2.5-7B-Instruct \citep{qwen2}            & 21.0           & 28.0               & 70.5              & 48.0              & 77.5             & 49.0               \\
    \quad + \ours                           & \textbf{26.0}  & \textbf{29.3}      & \textbf{72.5}     & \textbf{51.5}     & \textbf{79.5}    & \textbf{51.8}      \\
    \midrule
    Qwen-2.5-14B-Instruct \citep{qwen2}           & 21.0           & \textbf{32.0}      & 73.0              & 52.0              & 83.0             & 52.2               \\
    \quad + \ours                           & \textbf{24.0}  & 30.0               & \textbf{75.0}     & \textbf{57.0}     & \textbf{87.5}    & \textbf{54.7}      \\
    \midrule
    Llama-3.1-8B-Instruct \citep{dubey2024llama}  & 29.0           & 29.3               & 64.0              & 49.5              & 82.0             & 50.8               \\
    \quad + \ours                           & \textbf{32.5}  & \textbf{31.3}      & \textbf{68.0}     & \textbf{58.5}     & \textbf{84.5}    & \textbf{55.0}      \\
    \midrule
    Qwen-2.5-32B-Instruct \citep{qwen2}           & 24.5           & 26.0               & 72.0              & 55.0              & 88.0             & 53.1               \\
    Qwen-2.5-72B-Instruct \citep{qwen2}           & 27.0           & 28.7               & \textbf{74.5}     & 58.5              & \textbf{89.0}    & 55.5               \\
    Llama-3.1-70B-Instruct \citep{dubey2024llama} & \textbf{30.0}  & \textbf{33.3}      & 74.0              & \textbf{68.5}     & 85.5             & \textbf{58.3}      \\
    GPT-4o \citep{hurst2024gpt}                   & 21.5           & 28.0               & \textbf{74.5}     & 64.0              & 84.0             & 54.4               \\
\bottomrule
\end{tabular}
}
\caption{
Main evaluation results. Substring exact match (SubEM) serves as the evaluation metric, with the top-performing results emphasized in bold. \ours utilizes the training set of MuSiQue with self-supervision (\textsection\ref{sec:self_supervision}), and its performance on other tasks demonstrates the generalization ability of \ours.
}
\label{tab:main_results_models}
\end{table*}

\begin{table}[htbp]
\centering
\resizebox{\linewidth}{!}{
\begin{tabular}{lrrr}
\toprule
    Task                     & \# Example                & Max Tokens                & Avg. Tokens                \\ 
    \midrule
    Qasper                   & 200                       & 21,110                     & 4,921                     \\
    MultiFieldQA-en          & 150                       & 14,947                     & 6,888                     \\
    HotpotQA                 & 200                       & 16,322                     & 12,779                    \\
    MuSiQue                  & 200                       & 16,335                     & 15,542                    \\
    2WikiMultihopQA          & 200                       & 16,319                     & 7,096                     \\
    \bottomrule
\end{tabular}
}
\caption{
Statistics of evaluation tasks, with token counts calculated using the tokenizer of Llama-3.1-8B-Instruct.
}
\label{tab:task_statistics}
\end{table}

\begin{table}[htbp]
\centering
\resizebox{\linewidth}{!}{
\begin{tabular}{lccc}
\toprule
    Model                                                                   & Avg. Long-context               & Avg. Output Tokens           \\ 
    \midrule
    Qwen-2.5-Instruct 7B                                                    & 49.0                            & 375                     \\
    \quad + \ours                                                           & 51.8                            & 371                     \\
    Llama-3.1-Instruct 8B                                                   & 50.8                            & 289                     \\
    \quad + \ours                                                           & 55.0                            & 295                     \\
\bottomrule
\end{tabular}
}
\caption{
Average performance on long-context tasks (Tab. \ref{tab:main_results_models}) and average token count in model predictions for these tasks, measured with the model’s tokenizer.
}
\label{tab:output_tokens}
\end{table}

Leveraging self-provided supervision, we can either perform supervised fine-tuning on the highest-scoring outputs or apply preference optimization using preference pairs.

\paragraph{Supervised Fine-tuning.}
For supervised fine-tuning (SFT), we minimize the negative log-likelihood of the output as follows:
\begin{align}
    \mathcal{L}_{\text{SFT}} &= - \frac{1}{|y|} \log \pi_\theta(y \mid x) \nonumber \\
                             &= - \frac{1}{|y|} \sum_{i=1}^{|y|} \log \pi_\theta(y_i \mid x, y_{<i}) \nonumber 
\end{align}
Here, $y$ denotes the MBR decoding output.

\paragraph{Preference Optimization.}
Alternatively, we can conduct preference optimization to reinforce the tendency toward high-scoring outputs and reduce the likelihood of low-scoring outputs. Among the various preference optimization methods, we adopt the monolithic odds ratio preference optimization (ORPO) algorithm \citep{hong2024orpo} due to its strong empirical performance.  ORPO introduces an odds ratio loss to minimize the negative log odds ratio between a preferred output $y_w$ and a less-preferred output $y_l$:
\begin{align}
    \mathcal{L}_{\text{OR}} = -\log \sigma \left( \log \frac{\operatorname{odds}_\theta(y_w|x)}{\operatorname{odds}_\theta(y_l|x)} \right) \nonumber
\end{align}
Here, $\sigma$ represents the sigmoid function, and $\operatorname{odds}_\theta(y|x)$ measures how much more likely $y$ is to be generated than not:
\begin{align}
    \operatorname{odds}_\theta(y|x) = \frac{\pi_\theta(y|x)}{1-\pi_\theta(y|x)} \nonumber
\end{align}
The final objective in ORPO combines SFT and OR losses, with a hyperparameter $\beta$ controlling their relative importance:
\begin{align}
    \mathcal{L}_{\text{ORPO}} = \mathcal{L}_{\text{SFT}} + \beta \cdot \mathcal{L}_{\text{OR}} \nonumber
\end{align}
In our implementation, we use the MBR decoding output as $y_w$ and randomly select a low-scoring output to serve as $y_l$.
\section{Experiments}
\begin{table*}[htbp]
\centering
\resizebox{0.93\linewidth}{!}{
\begin{tabular}{lcccccc}
\toprule
    Model                                                            & Qasper         & MultiFieldQA-En        & HotpotQA         & MuSiQue           & 2WikiMQA         &  Avg.  \\
    \midrule
    Llama-3.1-8B-Instruct                                            & 29.0           & 29.3                   & 64.0             & 49.5              & 82.0             & 50.8    \\
    \midrule
    \multicolumn{7}{c}{\textit{Supervised Fine-tuning}} \\             
    \midrule
    \quad + TULU-V2-mix                                              & 26.5           & 27.3                   & 49.5             & 27.5              & 54.0             & 37.0     \\
    \quad + WildChat                                                 & 20.5           & 29.3                   & 46.5             & 28.0              & 58.0             & 36.5     \\
    \quad + LongAlpaca                                               & 22.5           & 31.3                   & 48.0             & 31.0              & 45.0             & 35.6     \\
    \quad + LongAlign                                                & 25.0           & \textbf{36.7}          & 58.5             & 47.5              & 76.0             & 48.7     \\
    \quad + LongMIT                                                  & 20.0           & 30.0                   & 56.0             & 36.0              & 66.5             & 41.7     \\
    \quad + LongReward-SFT                                           & 22.0           & 28.7                   & 58.0             & 52.0              & 76.5             & 47.4     \\
    \quad + GPT-4o-MuSiQue                                           & 21.5           & 31.3                   & 64.0             & \textbf{54.0}     & 83.5             & 50.9     \\
    \quad + \ours-SFT                                                & \textbf{28.5}  & 30.7                   & \textbf{68.5}    & 50.5              & \textbf{84.0}    & \textbf{52.4}     \\
    \midrule
    \multicolumn{7}{c}{\textit{Preference Optimization}} \\
    \midrule
    \quad + UltraFeedback                                            & 26.0           & 27.3                   & 47.5             & 28.5              & 46.0             & 35.1    \\ 
    \quad + LongReward-Preference                                    & 26.5           & \textbf{32.0}          & 63.5             & 52.0              & 80.5             & 50.9     \\
    \quad + \ours                                                    & \textbf{32.5}  & 31.3                   & \textbf{68.0}    & \textbf{58.5}     & \textbf{84.5}    & \textbf{55.0}     \\
\bottomrule
\end{tabular}
}
\caption{
A comparison between \ours and previous datasets. The results are based on Llama-3.1-8B-Instruct finetuned on the corresponding dataset. To ensure fairness, $2K$ examples are randomly sampled from each dataset, with the exception of TULU-V2-mix, WildChat, and UltraFeedback, where the longest $2K$ examples are selected. The preference optimization strategy is ORPO \citep{hong2024orpo}.
}
\label{tab:main_results_datasets}
\end{table*}

\begin{table}[t]
\centering
\resizebox{\linewidth}{!}{
\begin{tabular}{lcr}
\toprule
    Dataset                                               & Supervision            & Avg. Tokens                     \\ 
    \midrule
    TULU-V2-mix \citeyearpar{ivison2023camels}                  & [1], [2], [3]          & 3,788                          \\
    WildChat \citeyearpar{zhao2024wildchat}                     & [2], [3]               & 32,230                         \\
    LongAlpaca \citeyearpar{chen2024longlora}                   & [1], [4]               & 9,160                          \\             
    LongAlign  \citeyearpar{bai2024longalign}                   & [4]                    & 16,881                          \\
    LongMIT  \citeyearpar{chen2024essential}                    & [5]                    & 78,412                          \\
    LongReward-SFT \citeyearpar{zhang2024longreward}            & [6]                    & 22,206                          \\
    LongReward-Preference  \citeyearpar{zhang2024longreward}    & [6]                    & 22,689                          \\
    UltraFeedback \citeyearpar{cui2023ultrafeedback}            & [3]                    & 1,356                           \\
    GPT-4o-MuSiQue                                        & [7]                    & 18,476                           \\
    \ours                                                 & [8]                    & 18,532                           \\
\bottomrule
\end{tabular}
}
\caption{
Dataset statistics, including supervision source and average token count, measured with the Llama3.1-8B-Instruct tokenizer. Sources: [1] Human, [2] GPT-3.5-Turbo \citep{chatgpt}, [3] GPT-4 \citep{achiam2023gpt}, [4] Claude \citep{claude}, [5] Qwen2-72B-Instruct \citep{qwen2}, [6] GLM-4 \citep{glm2024chatglm}, [7] GPT-4o \citep{hurst2024gpt}, and [8] Self.
}
\label{tab:dataset_statistics}
\end{table}

\subsection{Implementation}
\ours requires query and long-context pairs to synthesize training data. Specifically, we leverage the training dataset of MuSiQue \citep{trivedi2022musique}, where each question is related to several Wikipedia documents. To achieve a specified number of tokens in the context, we randomly sample some unrelated documents, shuffle them with the related ones and concatenate them into a single context. We use the original questions in MuSiQue without the annotated answer, relying on the LLM to produce self-supervision (\textsection\ref{sec:self_supervision}). For each question, we sample $N=32$ outputs with a sampling temperature of $0.7$. By default, we synthesize $2048$ examples for fine-tuning, with context lengths randomly specified between $4K$ and $31K$ tokens. We conduct experiments using the Llama-3.1 models \citep{dubey2024llama} and Qwen-2.5 models \citep{qwen2}, with jina-embeddings-v3 serving as the sentence embedding model \citep{sturua2024jina}. ORPO \citep{hong2024orpo} is employed as the default fine-tuning method. More training details can be found in Appx. \ref{sec:training_details}.

\subsection{Evaluation Setup}
We conduct evaluations in long-context scenarios across a wide range of tasks. For single-document QA, we include Qasper \citep{dasigi2021dataset} and MultiFieldQA-En \citep{bai2023longbench} from the LongBench benchmark \citep{bai2023longbench}. For multi-document QA, we use HotpotQA \citep{yang2018hotpotqa}, MuSiQue \citep{trivedi2022musique} and 2WikiMultihopQA \citep{ho2020constructing}, also from LongBench. Task statistics are presented in Tab. \ref{tab:task_statistics}. We adopt plan-and-solve prompting for evaluation due to its strong performance (Tab. \ref{tab:prompting_methods}). Following previous research \citep{mallen2023not, asai2024selfrag, yen2024helmet}, we use substring exact match (SubEM) as the evaluation metric, measuring whether the output contains the golden answer.

\subsection{Main Results}
\paragraph{\ours Improves Various Models.}
We implement \ours on the leading open-source LLMs, including Qwen-2.5 models \citep{qwen2} and Llama-3.1 models \citep{dubey2024llama}. As illustrated in Tab. \ref{tab:main_results_models}, \ours brings notable improvements: when implemented on Qwen-2.5-7B-Instruct, it closes the performance gap with Qwen-2.5-14B-Instruct ($51.8$ vs. $52.2$); when applied to Qwen-2.5-14B-Instruct, it even exceeds the performance of Qwen-2.5-32B-Instruct ($54.7$ vs. $53.1$). Additionally, \ours yields an absolute improvement of $4.2$ on Llama-3.1-8B-Instruct, outperforming GPT-4o \citep{hurst2024gpt} ($55.0$ vs. $54.4$). Although \ours utilizes MuSiQue for data synthesis, it achieves strong performance across other tasks as well, highlighting its generalization potential. One possible shortcut of \ours is producing more tokens, as the evaluation metric, SubEM, might favor outputs with more tokens, which are more likely to contain the golden answer. To explore this, we examine output token counts. As shown in Tab. \ref{tab:output_tokens}, \ours has minimal effect on the number of output tokens.

\paragraph{\ours Competes with Previous Datasets.}

We compare \ours with several previous datasets, including short-context datasets such as TULU-V2-mix \citep{ivison2023camels}, WildChat \citep{zhao2024wildchat}, UltraFeedback \citep{cui2023ultrafeedback}, as well as long-context datasets including LongAlpaca \citep{chen2024longlora}, LongAlign \citep{bai2024longalign}, LongMIT \citep{chen2024essential}, and LongReward \citep{zhang2024longreward}. Additionally, we utilize GPT-4o to synthesize data using the same question and long-context as \ours, creating a dataset we term GPT-4o-MuSiQue. Dataset statistics are presented in Tab. \ref{tab:dataset_statistics}. To ensure fairness, $2K$ examples are randomly sampled from each dataset, with the exception of TULU-V2-mix, WildChat, and UltraFeedback, where the longest $2K$ examples are selected. As demonstrated in Tab. \ref{tab:main_results_datasets}, most previous datasets negatively affect the performance of Llama-3.1-8B-Instruct, consistent with the observation of \citet{gao2024train}. We hypothesize that this is because Llama-3.1-8B-Instruct already has strong long-context processing capabilities, and additional training on low-quality synthetic data could diminish its performance. However, we observe a performance improvement with \ours (50.8 to 55.0), indicating that self-improvement holds promise, which is particularlly promising as current LLMs advance rapidly.


\subsection{Analysis}
\label{sec:analysis}

\paragraph{Scoring Methods.}
\begin{table}[htbp]
\centering
\resizebox{\linewidth}{!}{
\begin{tabular}{lccc}
\toprule
    Method                                      & HotpotQA           & MuSiQue           & 2WikiMQA           \\ 
    \midrule
    Greedy Search                                & 64.0               & 49.5              & 82.0               \\
    \midrule
    Random                                       & 61.0               & 50.5              & 79.5               \\
    Reference-free Self-evaluation               & 64.0               & 51.5              & 83.0               \\
    \midrule
    \multicolumn{4}{c}{\textit{Minimum Bayes Risk}} \\
    \midrule
    ROUGE                                        & 66.5               & 53.5              & 85.0               \\
    BERTScore                                    & \textbf{67.5}      & 50.0              & 86.5               \\
    Reference-based Self-evaluation              & 63.5               & 51.5              & 84.5               \\
    Sentence Embedding                           & \textbf{67.5}      & \textbf{56.0}     & \textbf{88.0}      \\
\bottomrule
\end{tabular}
}
\caption{
Comparison of various scoring methods and greedy search. Each scoring method evaluates $16$ outputs sampled from Llama-3.1-8B-Instruct. The results indicate the performance of the highest-scoring output for each method.
}
\label{tab:scoring_methods}
\end{table}
Effective scoring methods are critical for creating self-supervision signals (\textsection\ref{sec:self_supervision}). We explore several approaches, including random selection, and reference-free self-evaluation, which prompts the LLM to assess its prediction in a separate turn based on the question and context. Additionally, we investigate various strategies for the utility metric $u(y, y^*)$ within Minimum Bayes Risk (\textsection\ref{sec:self_supervision}), such as ROUGE \citep{lin2004rouge}, BERTScore \citep{zhang2019bertscore} and reference-based self-evaluation, which prompts the LLM to assess $y$ using $y^*$ as the reference. The detailed prompts for reference-free and reference-based self-evaluation are presented in Tab. \ref{tab:self_eval_prompts} (Appx. \ref{sec:prompts}). For each question, we sample $N=16$ outputs using a temperature of $0.7$. Subsequently, we evaluate the highest-scoring output across different scoring methods and further compare these results to the performance of greedy search as a reference. As shown in Tab. \ref{tab:scoring_methods}, MBR-based methods outperform reference-free self-evaluation, even with simple N-gram-based ROUGE. We attribute this to the limited self-evaluation capabilities of current LLMs \citep{huang2024large, jiang2024self}, which might be more challenging in long-context scenarios. Incorporating more semantic information through sentence embeddings further improves MBR-based methods.

\paragraph{Number of Synthetic Examples.}
\begin{figure}[htbp]
\centering
\includegraphics[width=\linewidth]{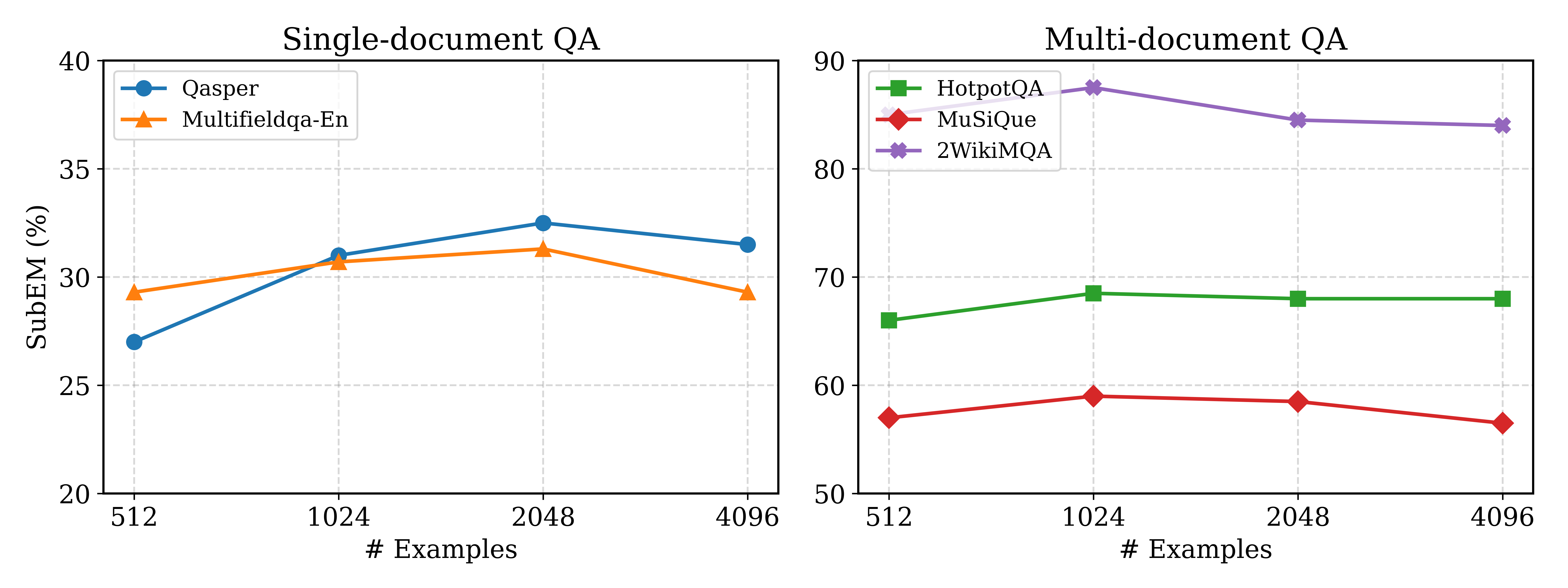}
\caption{
    Long-context performance of \ours with varying numbers of synthetic training examples, evaluated based on Llama-3.1-8B-Instruct fine-tuned on the corresponding dataset.
}
\label{fig:impact_of_the_number_of_training_examples}
\end{figure}
We analyze the impact of the number of training examples synthesized by \ours on long-context tasks. As shown in Fig. \ref{fig:impact_of_the_number_of_training_examples}, \ours demonstrates strong data efficiency, achieving competitive performance with only $1K$ examples, after which additional examples provide limited benefit. This suggests that \ours is unlocking the inherent potential of LLMs for long-context reasoning rather than introducing a new skill that would require more data.

\paragraph{Number of Samples per Example.}
\begin{figure}[htbp]
\centering
\includegraphics[width=\linewidth]{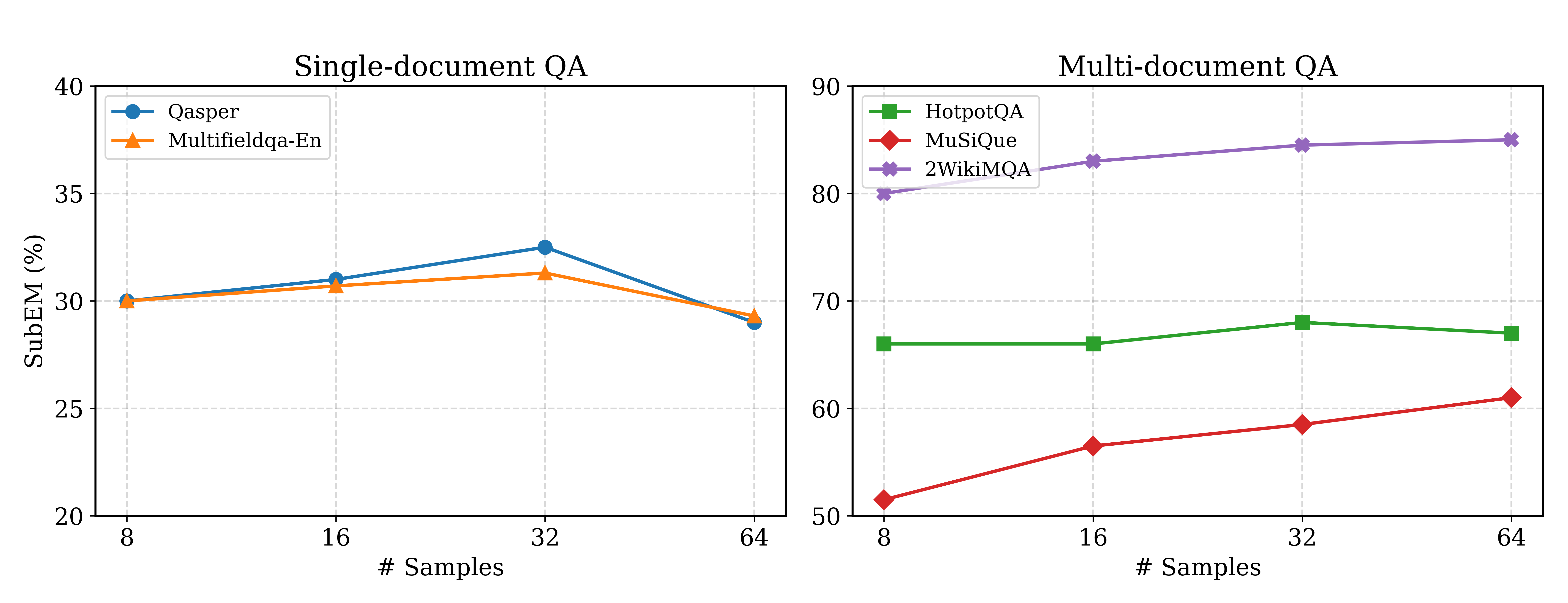}
\caption{
    Long-context performance of \ours with varying numbers of samples per example during data synthesis, evaluated based on Llama-3.1-8B-Instruct fine-tuned on the corresponding dataset.
}
\label{fig:impact_of_the_number_of_samples_per_example}
\end{figure}
We continue to explore the effect of the number of samples, $N$, per example. As illustrated in Fig. \ref{fig:impact_of_the_number_of_samples_per_example}, increasing $N$ from $8$ to $32$ consistently improves performance, likely due to more accurate MBR estimation (\textsection\ref{sec:self_supervision}). Beyond $32$, except for MuSiQue, the performance improvement diminishes. This may indicate a fundamental limitation of our scoring method, which appears to struggle with selecting higher-quality outputs from larger output sets when $N > 32$ (see also in Tab. \ref{figure:llms_potential_for_correct_long_context_reasoning}). We believe the scoring method is pivotal to self-improvement and will investigate this aspect further in future work.

\paragraph{Short-context Performance.}
\begin{table*}[htbp]
\centering
\resizebox{\linewidth}{!}{
\begin{tabular}{lcccccccc}
\toprule
    \multirow{2}{*}{Model} & \multicolumn{1}{c}{Long-Context}         & \multicolumn{7}{c}{Short-Context}        \\
                             \cmidrule(lr){2-2}                                           \cmidrule(lr){3-9}                                          
                              & Avg.            & MMLU         & GSM8K           & ARC-Challenge       & HellaSwag        & Winogrande        & TruthfulQA         & Avg.        \\
    \midrule
    Qwen-2.5-7B-Instruct      & 49.0            & 74.2         & 82.4            & 67.1        & 81.5             & 74.7              & 64.7               & 74.1               \\
    \quad + \ours             & \textbf{51.8}            & 74.1         & 83.2            & 66.5        & 81.3             & 74.4              & 64.8               & 74.1               \\
    \midrule
    Llama-3.1-8B-Instruct     & 50.8            & 68.3         & 77.7            & 60.2        & 80.1             & 77.4              & 54.1               & 69.6               \\
    \quad + \ours       & \textbf{55.0}            & 68.4         & 77.8            & 60.3        & 79.9             & 77.3              & 53.8               & 69.6             \\
\bottomrule
\end{tabular}
}
\caption{
Evaluation results on short-context tasks from the Open LLM Leaderboard \citep{open-llm-leaderboard}, with the long-context average performance referenced from Tab.\ref{tab:main_results_models}. \ours demonstrates a marked improvement in long-context performance, with minimal impact on short-context performance.
}
\label{tab:short_context_performance}
\end{table*}
Improving long-context reasoning should not compromise short-context performance. To investigate this further, we evaluate \ours on the Open LLM Leaderboard \citep{open-llm-leaderboard}, covering $6$ tasks that represent diverse capabilities: MMLU \citep{hendrycks2021measuring}, GSM8K \citep{cobbe2021training}, ARC-challenge \citep{clark2018think}, HellaSwag \citep{zellers2019hellaswag}, WinoGrande \citep{sakaguchi2021winogrande} and TruthfulQA \citep{lin2022truthfulqa}. As shown in Tab. \ref{tab:short_context_performance}, while \ours achieves substantial improvements in long-context performance, it has minimal impact on short-context performance. 

\section{Related Work}

\paragraph{Long-context Language Modeling.}
Numerous studies explore methods to extend the long-context processing abilities of LLMs. One line of research approaches addresses this challenge from a model-centered perspective, with some studies focusing on minimal modifications to existing LLMs, such as adjustments to position embeddings \citep{chen2023extending, peng2024yarn, ding2024longrope, zhu2024pose, xiong-etal-2024-effective} and refinements to the attention mechanism \citep{ding2023longnet, jin2024llm, an2024trainingfree, an2024does}.
Additionally, some works propose novel architectures for efficient long-context processing \citep{wu2022memorizing, bertsch2024unlimiformer, wang2024augmenting, yen2024long, lieber2024jamba, ye2024differential, sun2024you}. Another line of research adopts a data-centric perspective, focusing on data engineering strategies. For example, \citet{dubey2024llama, lieber2024jamba, fu2024data, gao2024train} continue pre-training models on long sequences, while \citet{an2024make, bai2024longalign, zhang2024longreward, chen2024essential, chen2024longlora} leverage expert models or human annotations to create long-context data for fine-tuning. In contrast to these approaches, this work aims to facilitate the self-improvement of LLMs in long-context reasoning.

\paragraph{Self-improving.}
The self-improvement of LLMs has become a vital area of research as these models advance toward human intelligence. Research in this area follows two main approaches. The first approach investigates the self-reflection capabilities of LLMs, where models are prompted to assess and refine their own outputs \citep{ganguli2023capacity, madaan2024self, shinn2024reflexion, xie2024self, gou2024critic, chen2024teaching, pan2024automatically}. However, the reliability of these self-refinement has been questioned in recent studies \citep{huang2024large, jiang2024self}. The second approach involves generating synthetic training data through the models themselves. This process typically involves generating multiple outputs for a given input, filtering out inaccurate results based on ground-truths, and using the remaining correct responses for model fine-tuning \citet{zelikman2022star, hosseini2024vstar, pang2024iterative, wang2024self, gulcehre2023reinforced, zhang2024rest}. Additionally, \citet{yuan2024selfrewarding} fine-tune LLMs to assign rewards to their own outputs using human preference data and facilitate continual improvement in instruction following. To reduce reliance on human annotations, some studies adopts consensus-based supervision, designating the output with the higher consensus across multiple outputs as better, with applications in areas such as arithmetic and logical reasoning \citep{huang2023large, prasad2024selfconsistencypreferenceoptimization}, machine translation \citep{finkelstein2024mbr, wang2024don, yang2024direct}, and instruction following \citep{wu2024better}. \ours first reveals the underestimated potential of LLMs in long-context reasoning and then leverages a consensus-based supervision strategy to enable LLMs to self-improve in long-context reasoning.

\section{Conclusion}
In this study, we investigate the potential of LLMs to self-improve in long-context reasoning and propose \ours for this purpose. This method achieves substantial improvements across multiple long-context reasoning tasks. We hope this research will open new avenues for self-improvement in long-context reasoning, which is vital for the sustained progress of LLMs, particularly as they advance toward surpassing human intelligence.
\section*{Limitations}

We recognize that this work has several limitations that warrant further investigation.

\paragraph{Scoring Method.}
To establish self-supervision (\textsection\ref{sec:self_supervision}), we score each output according to Minimum Bayesian Risk (MBR), which reflects consensus across multiple sampled outputs. However, a substantial performance gap remains between the highest MBR-scored output and the oracle sample (see Tab. \ref{figure:llms_potential_for_correct_long_context_reasoning} for details). Future research should explore more effective approaches for self-evaluation of outputs. One possible direction could involve examining the critic capabilities of LLMs in long-context scenarios \citep{lan2024criticbench, lin-etal-2024-criticbench, lan2024training}.

\paragraph{Synthetic Data.}
Another limitation of this work is its reliance on MuSiQue \citep{trivedi2022musique} for synthetic data, which consists of multi-hop questions spanning multiple paragraphs. While this approach has enabled some progress, MuSiQue dose not cover all challenging question types, such as those requiring full-context reasoning, which remains a key limitation of current long-context LLMs \citep{karpinska2024one, wang2024leave, vodrahalli2024michelangelo, yen2024helmet}. We advocate for future work to prioritize the creation of high-quality prompt sets, which are essential for the development of long-context LLMs.

\paragraph{Experimental Setup.} 
Due to the computational limitations, we restrict the implementation of \ours to LLMs with up to $14$B parameters, though its effectiveness at larger scales warrants further investigation. Likewise, the maximum sequence length is set to $32K$ tokens, whereas current leading LLMs support context lengths of up to $128K$ tokens or more. We leave the exploration of longer context lengths for future work.

\bibliography{custom}

\clearpage

\appendix

\section{Training Details}
\label{sec:training_details}

To support efficient fine-tuning for long-context scenarios, we implement sequence parallelization \citep{jacobs2023deepspeed} with a parallel size of 8. Additionally, we utilize QLoRA \citep{dettmers2024qlora} to reduce memory consumption during fine-tuning. The LoRA rank, alpha, and dropout are set to $128$, $128$, and $0.05$, respectively, with all attention and feedforward linear layers designated as target modules. All models are fine-tuned for one epoch. The batch size, learning rate, and maximum sequence length are set to $8$, $5e-5$, and $32K$, respectively. The $\beta$ for ORPO is configured to $0.1$. All experiments are conducted on a computing setup with $8 \times \text{H}100$ GPUs.

\section{Prompts}
\label{sec:prompts}

\begin{table*}[htbp]
\centering
\begin{tabular}{l|p{0.72\linewidth}}
\toprule
    \textbf{Strategy} & \textbf{Prompt} \\
\midrule
    \multirow{3}{*}{Default} & \textnormal{\{context\}} \\ 
                             & \textnormal{} \\ 
                             & \textnormal{\{input\}} \\ 
    \midrule
    \multirow{4}{*}{Direct Answer} & \textnormal{\{context\}} \\  
                                    & \textnormal{} \\ 
                                    & \textnormal{\{input\}} \\  
                                    & \textnormal{Let's answer the question directly.} \\  
    \midrule
    \multirow{4}{*}{\makecell[l]{Think step-by-step \\ \citep{kojima2022large}}} & \textnormal{\{context\}} \\  
                                                                                 & \textnormal{} \\
                                                                                 & \textnormal{\{input\}} \\  
                                                                                 & \textnormal{Let's think step by step.} \\  
    \midrule
    \multirow{4}{*}{\makecell[l]{Fact-and-reflection \\ \citep{zhao-etal-2024-fact}}} & \textnormal{\{context\}} \\  
                                                                                      & \textnormal{} \\ 
                                                                                      & \textnormal{\{input\}} \\  
                                                                                      & \textnormal{Let's first identify the relevant information from the long context and list it. Then, carry out step-by-step reasoning based on that information, and finally, provide the answer.} \\  
    \midrule
    \multirow{4}{*}{\makecell[l]{Plan-and-solve \\ \citep{wang2023plan}}} & \textnormal{\{context\}} \\  
                                                                          & \textnormal{} \\ 
                                                                          & \textnormal{\{input\}} \\  
                                                                          & \textnormal{Let's first understand the problem and devise a plan to solve it. Then, let's carry out the plan and solve the problem step-by-step.} \\  
\bottomrule
\end{tabular}
\caption{The prompts for various prompting strategies (\textsection\ref{sec:prompting_strategies_matter}), where \{context\} and \{input\} serve as placeholders for the long context and input query, respectively.}
\label{tab:prompts}
\end{table*}

\begin{table*}[htbp]
\centering
\begin{tabular}{l|p{0.72\linewidth}}
\toprule
    \textbf{Strategy} & \textbf{Prompt} \\
\midrule
    \multirow{10}{*}{\makecell[l]{Reference-free \\ Self-Evaluation}} & \textnormal{[Context]} \\  
                                                      & \textnormal{\{context\}} \\  
                                                      & \textnormal{} \\ 
                                                      & \textnormal{[Question]} \\  
                                                      & \textnormal{\{question\}} \\  
                                                      & \textnormal{} \\ 
                                                      & \textnormal{[Predicted Response]} \\  
                                                      & \textnormal{\{prediction\}} \\  
                                                      & \textnormal{} \\ 
                                                      & \textnormal{Please evaluate the correctness of the predicted response based on the context and the question. Begin your evaluation by providing a brief explanation. Be as objective as possible. After giving your explanation, you must rate the response on a scale from 1 to 5, following this format exactly: “[[rating]]”. For example, “Rating: [[3]]”.} \\  
    \midrule
    \multirow{10}{*}{\makecell[l]{Reference-based \\ Self-Evaluation}} & \textnormal{Here is a question along with two responses: one is the reference response, and the other is the predicted response. Please determine whether the two responses provide the same answer to the question. Respond with “True” or “False” directly.} \\ 
                                                     & \textnormal{} \\ 
                                                     & \textnormal{[Question]} \\  
                                                     & \textnormal{\{question\}} \\ 
                                                     & \textnormal{} \\ 
                                                     & \textnormal{[Reference Response]} \\ 
                                                     & \textnormal{\{reference\}} \\ 
                                                     & \textnormal{} \\ 
                                                     & \textnormal{[Predicted Response]} \\ 
                                                     & \textnormal{\{prediction\}} \\ 
\bottomrule
\end{tabular}
\caption{The prompts for the reference-free and reference-based self-evaluation strategies (\textsection\ref{sec:analysis}), where \{question\}, \{reference\}, \{prediction\}, and \{context\} serve as placeholders for their respective elements.}
\label{tab:self_eval_prompts}
\end{table*}

We provide the prompts for various prompting strategies (\textsection\ref{sec:prompting_strategies_matter}) in Tab. \ref{tab:prompts}, and the prompts for the reference-free and reference-based self-evaluation strategies (\textsection\ref{sec:analysis}) in Tab. \ref{tab:self_eval_prompts}.

\end{document}